\documentclass{article} %
\usepackage{iclr2019_conference,times}
\usepackage{amsmath,amsfonts,bm}

\def\eqref#1{equation~\ref{#1}}

\def\1{\bm{1}}

\def\va{{\bm{a}}}

\def\ve{{\bm{e}}}

\def\vh{{\bm{h}}}

\def\vp{{\bm{p}}}

\def\vx{{\bm{x}}}

\def\vz{{\bm{z}}}

\def\mA{{\bm{A}}}

\def\mH{{\bm{H}}}

\def\mM{{\bm{M}}}

\def\mW{{\bm{W}}}
\def\mX{{\bm{X}}}

\def\mZ{{\bm{Z}}}

\DeclareMathAlphabet{\mathsfit}{\encodingdefault}{\sfdefault}{m}{sl}
\SetMathAlphabet{\mathsfit}{bold}{\encodingdefault}{\sfdefault}{bx}{n}

\def\gE{{\mathcal{E}}}

\def\gG{{\mathcal{G}}}

\def\gV{{\mathcal{V}}}

\def\sG{{\mathbb{G}}}

\def\sR{{\mathbb{R}}}

\usepackage{hyperref}
\usepackage{url}
\usepackage{graphicx}
\usepackage{booktabs}
\usepackage{multirow}
\usepackage{color, colortbl}
\usepackage{subfigure}
\usepackage{enumitem}
\title{Dynamic Graph Representation Learning via Self-Attention Networks}

\author{Aravind Sankar\thanks{Work done while at Visa Research} \\
Department of Computer Science\\
University of Illinois, Urbana-Champaign\\
\texttt{asankar3@illinois.edu} \\
\And
Yanhong Wu, Liang Gou, Wei Zhang \& Hao Yang\\
Visa Research \\
Palo Alto, CA\\
\texttt{\{yanwu,ligou,wzhan,haoyang‎\}@visa.com} \\
}

\definecolor{darkred}{RGB}{168,37,17}
\definecolor{darkblue}{RGB}{23,55,119}
\definecolor{highblue}{RGB}{20, 20, 180}
\definecolor{lightblue}{RGB}{183,210,237}
\definecolor{gray}{RGB}{100,100,100}
\definecolor{lightgray}{RGB}{230,230,230}
\definecolor{rainforest}{RGB}{3,101,100}
\definecolor{darkpurple}{RGB}{66,8,91}
\definecolor{orange}{RGB}{242, 101, 34}
\definecolor{black}{RGB}{0, 0, 0}

\newcommand{\name}{DySAT}

\newcommand{\add}[1]{\textcolor{blue}{#1}}

\renewcommand{\add}[1]{#1}

\newcolumntype{K}[1]{>{\centering\arraybackslash}p{#1}}

\iclrfinalcopy %
\begin{document}

\maketitle

\begin{abstract}
\label{sec:abstract}
Learning latent representations of nodes in graphs is an important and ubiquitous task with widespread applications such as link prediction, node classification, and graph visualization.
Previous methods on graph representation learning mainly focus on static graphs, however, many real-world graphs are dynamic and evolve over time.
In this paper, we present Dynamic Self-Attention Network (\name), a novel neural architecture that operates on dynamic graphs and learns node representations that capture both structural properties and temporal evolutionary patterns.
Specifically, \name~computes node representations by jointly employing self-attention layers along two dimensions: structural neighborhood and temporal dynamics.
We conduct link prediction experiments on two classes of graphs: communication networks and bipartite rating networks.
Our experimental results show that~\name~has a significant performance gain over several different state-of-the-art graph embedding baselines.
\end{abstract}

\section{Introduction}
\label{sec:introduction}
Learning latent representations (or embeddings) of nodes in graphs has been recognized as a fundamental learning problem due to its widespread use in various domains such as social media~\citep{deepwalk}, biology~\citep{node2vec}, and knowledge bases~\citep{knowledge_embedding}.
The basic idea is to learn a low-dimensional vector for each node, which encodes the structural properties of a node and its neighborhood (and possibly attributes).
Such low-dimensional representations can benefit a plethora of graph analytical tasks such as node classification, link prediction, recommendation and graph visualization~\citep{deepwalk, line,node2vec,sdne,pinsage,adv18}.

Previous work on graph representation learning mainly focuses on static graphs, which contain a fixed set of nodes and edges.
However, many graphs in real-world applications are intrinsically dynamic, in which graph structures can evolve over time.
They are usually represented as a sequence of graph snapshots from different time steps~\citep{dynamic_snapshot}.
Examples include academic co-authorship networks where authors may periodically switch their collaboration behaviors and email communication networks whose structures may change dramatically due to sudden events.
In such scenarios, modeling temporal evolutionary patterns is important in accurately predicting node properties and future links.

Learning dynamic node representations is challenging, compared to static settings, due to the complex time-varying graph structures: nodes can emerge and leave, links can appear and disappear, and communities can merge and split.
This requires the learned embeddings not only to preserve structural proximity of nodes, but also to jointly capture the temporal dependencies over time.
Though some recent work attempts to learn node representations in dynamic graphs, they mainly impose a temporal regularizer to enforce smoothness of the node representations from adjacent snapshots ~\citep{smoothness,attributes_cikm17,dynamictriad}.
However, these approaches may fail when 
nodes exhibit significantly distinct evolutionary behaviors.
\citet{know-evolve} employ a recurrent neural architecture for temporal reasoning in multi-relational knowledge graphs. However, their temporal node representations are limited to modeling first-order proximity, while ignoring the structure of higher-order graph neighborhoods.

Attention mechanisms have recently achieved great success in many sequential learning tasks such as machine translation~\citep{bahdanau} and reading comprehension~\citep{qanet}.
The key underlying principle is to learn a function that aggregates a variable-sized input, while focusing on the parts most relevant to a certain context.
When the attention mechanism uses a single sequence as both the inputs and the context, it is often called \textit{self-attention}.
Though attention mechanisms were initially designed to facilitate Recurrent Neural Networks (RNNs) to capture long-term dependencies, recent work by~\citet{self-attention} demonstrates that a fully self-attentional network itself can achieve state-of-the-art performance in machine translation tasks.
~\citet{gat} extend self-attention to graphs by enabling each node to attend over its neighbors, achieving state-of-the-art results for semi-supervised node classification tasks in static graphs. 

As dynamic graphs usually include periodical patterns such as recurrent links or communities, attention mechanisms are capable of utilizing information about most relevant historical context, to facilitate future prediction. 
Inspired by recent work on attention techniques, we present a novel neural architecture named Dynamic Self-Attention Network (\name) to learn node representations on dynamic graphs.
Specifically, we employ self-attention along two dimensions: structural neighborhoods and temporal dynamics, \textit{i.e.},~\name~generates a dynamic representation for a node by considering both its neighbors and historical representations, following a self-attentional strategy.
Unlike static graph embedding methods that focus entirely on preserving structural proximity, we learn dynamic node representations that reflect the temporal evolution of graph structure over a varying number of historical snapshots.
In contrast to temporal smoothness-based methods,~\name~learns attention weights that capture temporal dependencies at a fine-grained node-level granularity.

 We evaluate our framework on the dynamic link prediction task using four benchmarks of different sizes including two email communication networks~\citep{enron,UCI} and two bipartite rating networks~\citep{ml-10m}.
 Our evaluation results show that~\name~achieves significant improvements (3.6\% macro-AUC on average) over several state-of-the-art baselines and maintains a more stable performance over different time steps.

\section{Related Work}
\label{sec:related_work}
 
Our framework is related to previous representation learning techniques on static graphs,  dynamic graphs, and recent developments in self-attention mechanisms.

\textbf{Static graph embeddings.}
Early work on unsupervised graph representation learning exploits the spectral properties of various graph matrix representations, such as Laplacian, etc. to perform dimensionality reduction~\citep{dimension_reduction,eigenmap}.
To improve scalability, some work~\citep{deepwalk,node2vec} utilizes Skip-gram methods, inspired by their success in Natural Language Processing (NLP).
Recently, several graph neural network architectures based on generalizations of convolutions have achieved tremendous success, among which many methods are designed for supervised or semi-supervised learning tasks~\citep{pscn,cheby,gcn,motif,gat}.
~\citet{graphsage} extend graph convolutional methods through trainable neighborhood aggregation functions, to propose a general framework applicable to unsupervised representation learning.
However, these methods are not designed to model temporal evolutionary patterns in dynamic graphs.

\textbf{Dynamic graph embeddings.}
Most techniques employ temporal smoothness regularization to ensure embedding stability across consecutive time-steps~\citep{smoothness}.
~\citet{dynamictriad} additionally use triadic closure~\citep{triadic} as guidance, leading to significant improvements.
Neural methods were recently explored in the knowledge graph domain by~\citet{know-evolve}, who employ a recurrent neural architecture for temporal reasoning.
However, their model is limited to tracing link evolution, thus limited to capturing first-order proximity.
~\citet{dyngem} learn incremental node embeddings through initialization from the previous time steps, however, this may not guarantee the model to capture long-term graph similarity.
\add{A few recent works~\citep{ctdne,kdd_cont,dyrep,streamnn,jodie,latent} examine a related setting of temporal graphs with continuous time-stamped links for representation learning, which is however orthogonal to the established problem setup of using dynamic graph snapshots.
\citet{attributes_cikm17} learn node embeddings in dynamic attributed graphs by initially training an offline model, followed by incremental updates over time. However, their key focus is online learning to improve efficiency over re-training static models, while our goal is to improve  representation quality by exploiting the temporal evolutionary patterns in graph structure.}
Unlike previous approaches, our framework captures the most relevant historical contexts through a self-attentional architecture, to learn dynamic node representations. 

\textbf{Self-attention mechanisms.}
Recent advancements in many NLP tasks have demonstrated the superiority of \emph{self-attention} in achieving state-of-the-art performance~\citep{self-attention,structured_attention, role_labeling, disan, relative_position}. 
In ~\name, we employ self-attention mechanisms to compute a dynamic node representation by attending over its neighbors and previous historical representations. 
Our approach of using self-attention over neighbors is closely related to the Graph Attention Network (GAT)~\citep{gat}, which employs neighborhood attention for semi-supervised node classification in a static graph.
As dynamic graphs usually contain periodical patterns, we extend the self-attention mechanisms over the historical representations of a particular node to capture its temporal evolution behaviors.

\section{Problem Definition}
In this work, we address the problem of dynamic graph representation learning. A dynamic graph is defined as a series of observed snapshots, $ \sG = \{\gG^1, \dots, \gG^T \}$ where $T$ is the number of time steps.
Each snapshot $\gG_t = (\gV, \gE^t)$ is a weighted undirected graph with a shared node set $\gV$, a link set $\gE^t$, and weighted adjacency matrix $\mA^t$ at time $t$.
Unlike some previous work that assumes links can only be added over time in dynamic works, we also allow to remove links.
Dynamic graph representation learning aims to learn latent representations $\ve^t_v \in \sR^d$ for each node $v \in \gV$ at time steps $t =1, 2, \dots, T$, such that $\ve^t_v$ preserves both the local graph structures centered at $v$ and its evolutionary behaviors prior to time $t$.

\section{Dynamic Self-Attention Network}
In this section, we first describe the high-level structure of our model. \name~consists of two major novel components: \emph{structural} and \emph{temporal} \emph{self-attention} layers, which can be utilized to construct arbitrary graph neural architectures through stacking of layers.
Similar to existing studies on attention mechanisms, we employ multi-head attentions to improve model capacity and stability.

\name~ consists of a \emph{structural} block followed by a \emph{temporal} block, as illustrated in Figure~\ref{fig:dysat}, where each block may contain multiple stacked layers of the corresponding layer type. 
The \emph{structural} block extracts features from the local neighborhood through self-attentional aggregation, to compute intermediate node representations for each snapshot. 
These representations feed as input to the \emph{temporal} block, which attends over multiple time steps, capturing temporal variations in the graph.

\subsection{Structural self-attention}
The input of this layer is a graph snapshot $\gG \in \sG$ and a set of input node representations 
$\{ \vx_v \in \sR^D, \forall v \in \gV \}$ where $D$ is the input embedding dimension. 
The input to the initial layer can be set as 1-hot encoded vectors for each node (or attributes if available). 
The output is a new set of node representations $\{\vz_v \in \sR^F, \forall v \in \gV \}$ with  $F$ dimensions that capture local structural properties.

Specifically, the \emph{structural} self-attention layer attends over the immediate neighbors of a node $v$ (in snapshot $\gG$), by computing attention weights as a function of their input node embeddings.
The structural attention layer is a variant of GAT~\citep{gat}, applied on a single snapshot:
\begin{equation}
 \vz_v = \sigma \Big( \sum\limits_{u \in \mathcal{N}_v} \alpha_{uv} \mW^s \vx_u \Big), \hspace{10pt} \alpha_{uv} = \frac{\exp \Big(\sigma  \Big(A_{uv} \cdot \va^T [ \mW^s \vx_u || \mW^s \vx_v ]\Big)\Big)}{ \sum\limits_{w \in \mathcal{N}_v} \exp \Big(\sigma\Big(A_{wv} \cdot \va^T [ \mW^s \vx_w || \mW^s \vx_v ]\Big)\Big) }
 \vspace{-5pt}
\end{equation}

where $\mathcal{N}_v = \{ u \in \gV : (u, v) \in \gE \}$ is the set of immediate neighbors of node $v$ in snapshot $\gG$;
$\mW^s \in \sR^{D \times F}$ is a shared weight transformation applied to each node in the graph;
$\va \in \sR^{2D}$ is a weight vector parameterizing the attention function implemented as feed-forward layer;
$||$ is the concatenation operation and $\sigma(\cdot)$ is a non-linear activation function.
Note that $A_{uv}$ is the weight of link $(u,v)$ in the current snapshot $\gG$.
The set of learned coefficients $\alpha_{uv}$, obtained by a softmax over the neighbors of each node, indicate the importance or contribution of node $u$ to node $v$ at the current snapshot. 
We use a LeakyRELU non-linearity to compute the attention weights, followed by ELU for the output representations. In our experiments, we employ sparse matrices to implement the \emph{masked} self-attention over neighbors. %

\subsection{Temporal self-attention}
To further capture temporal evolutionary patterns in a dynamic network, we design a temporal self-attention layer.
The input of this layer is a sequence of representations of a particular node $v$
at different time steps.
Specifically, for each node $v$, we define the input as $\{ \vx^1_v, \vx^2_v, \dots, \vx^T_v\}, \vx^t_v \in \sR^{D'}$ where $T$ is the number of time steps and 
$D'$ is the dimensionality of the input representations.	
The layer output is a new representation sequence for $v$ at each time step, \textit{i.e.}, $\vz_v = \{\vz^1_v, \vz^2_v, \dots, \vz^T_v \}, \vz^t_v \in \sR^{F'}$ with dimensionality $F'$. 
We denote the input and output representations of $v$, packed together across time, by matrices $\mX_v \in \sR^{T \times D'}$ and $\mZ_v \in \sR^{T \times F'}$ respectively. 

The key objective of the temporal self-attentional layer is to capture the temporal variations in graph structure over multiple time steps.
The input representation of node $v$ at time-step $t$, $\vx^t_v$, constitutes an encoding of the current local structure around $v$.
We use $\vx^t_v$ as the query to attend over its historical representations ($<t$), tracing the evolution of the local neighborhood around $v$.
Thus, temporal self-attention facilitates learning of dependencies between various representations of a node across different time steps.

To compute the output representation of node $v$ at $t$, we use the scaled dot-product form of attention~\citep{self-attention} where the queries, keys, and values are set as the input node representations.
The queries, keys, and values are first transformed to a different space by using linear projections matrices $\mW_q \in \sR^{D' \times F'}, \mW_k \in \sR^{D' \times F'}$ and $\mW_v \in \sR^{D' \times F'}$ respectively.
Here, we allow each time-step $t$ to attend over all time-steps up to and including $t$, to prevent leftward information flow and preserve the auto-regressive property.
The temporal self-attention is defined as:
\vspace{-5pt}
\begin{equation}
\mZ_v = \bm{\beta_v} (\mX_v \mW_v), \hspace{10pt} \beta^{ij}_v = \frac{\exp(e^{ij}_v)}{\sum\limits_{k=1}^T \exp(e^{ik}_v)}, \hspace{10pt} e^{ij}_v = \Big(\frac{((\mX_v \mW_q) (\mX_v \mW_k )^T)_{ij} }{\sqrt{F'}} + M_{ij} \Big)
\vspace{-5pt}
\end{equation}

where $\bm{\beta_v} \in \sR^{T \times T}$ is the attention weight matrix obtained by the multiplicative attention function and $\mM \in \sR^{T \times T} $ is a mask matrix with each entry $M_{ij} \in \{-\infty, 0\}$.
When $M_{ij} = -\infty$, the softmax function results in a zero attention weight, \textit{i.e.},  $\beta_v^{ij} = 0$, which switches off the attention from time-step $i$ to $j$. To encode the temporal order, we define $\mM$ as:
\[ M_{ij} = 
\begin{cases}
 0, & i \leq j \\
- \infty, &  \text{otherwise}  \\
\end{cases}
\]
\begin{figure}[t]
\includegraphics[width=\linewidth]{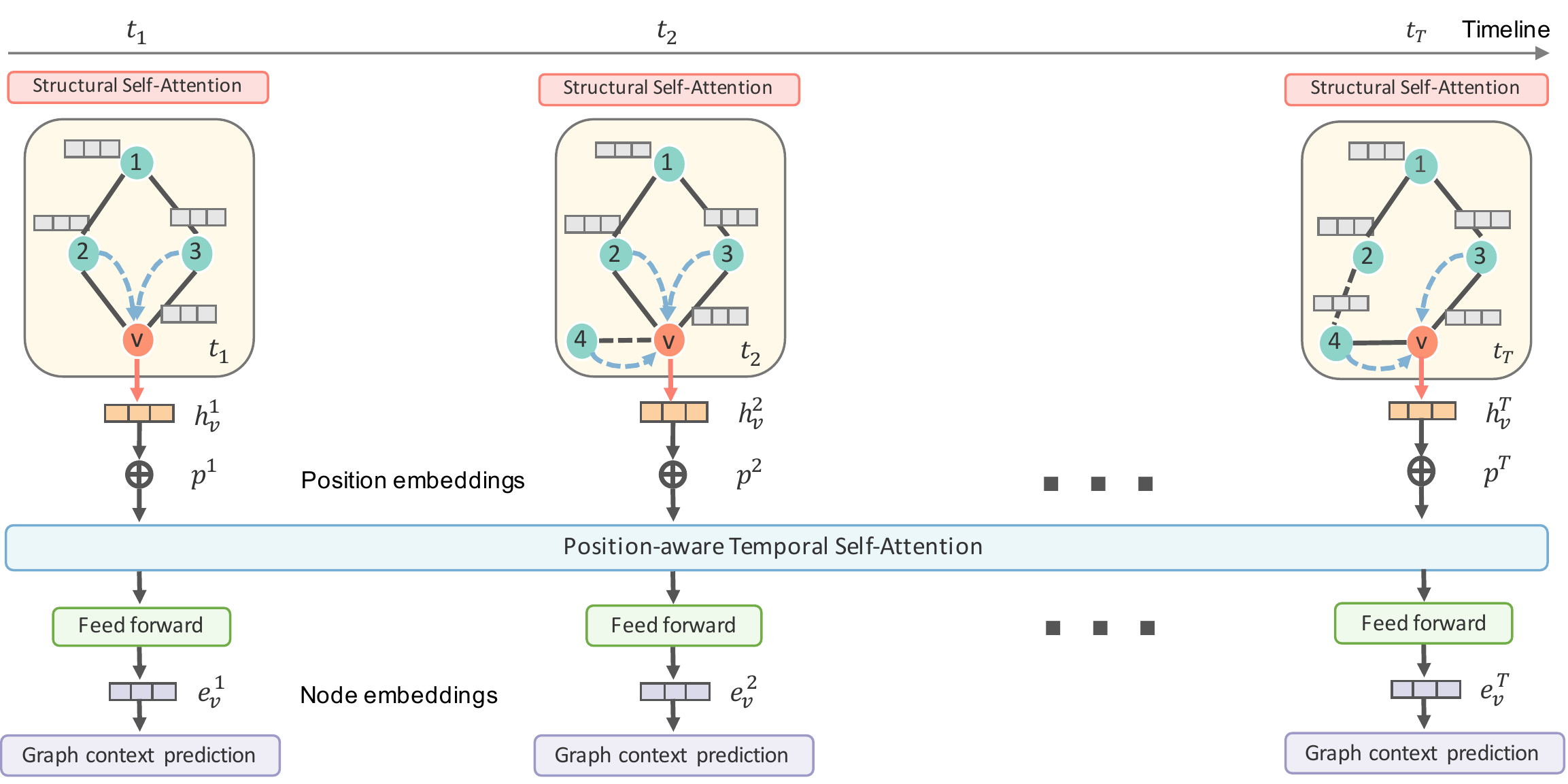}
\vspace{-15pt}
\caption{Neural architecture of~\name: we employ structural attention layers followed by temporal attention layers. Dashed black arrows indicate new links and dashed blue arrows refer to neighbor-based structural-attention.}
\label{fig:dysat}
\end{figure}

\subsection{Multi-Head Attention}
We additionally employ multi-head attention~\citep{self-attention} to jointly attend to different subspaces at each input, leading to a leap in model capacity.
We use multiple attention heads, followed by concatenation, in both structural and temporal self-attention layers: 
\begin{align}
\text{Structural multi-head self-attention:} & \hspace{30pt} \vh_v = \text{Concat} (\vz_v^1, \vz_v^2, \dots, \vz_v^H) \; \; & \forall v \in V \\ 
\text{Temporal multi-head self-attention:} & \hspace{30pt} \mH_v = \text{Concat} (\mZ_v^1, \mZ_v^2,  \dots, \mZ_v^H) \; \;  & \forall v \in V
\vspace{-10pt}
\end{align}

where $H$ is the number of attention heads, $\vh_v \in \sR^F$ and $\mH_v \in \sR^{T \times F'}$ are the outputs of structural and temporal multi-head attentions respectively.
Note that while structural attention is applied on a single snapshot, temporal attention operates over multiple time-steps.

\subsection{DySAT Architecture}
In this section, we present our neural architecture DySAT for Dynamic Graph Representation Learning, that uses the above defined \emph{structural} and \emph{temporal} self-attention layers as fundamental modules. 
As shown in Figure~\ref{fig:dysat}, DySAT has three modules from its top to bottom, (1) \emph{structural} attention block, (2) \emph{temporal} attention block, and (3) 
graph context prediction.
The model takes as input a collection of $T$ graph snapshots, and generates outputs latent node representations at each time step.

\textbf{Structural attention block.}
This module is composed of multiple stacked structural self-attention layers to extract features from nodes at different distances.
We apply each layer independently at different snapshots with shared parameters, as illustrated in Figure~\ref{fig:dysat}, to capture local neighborhood structure around a node at each time step. 
Note that the embeddings input to a layer can potentially vary across different snapshots.
We denote the node representations output by the structural attention block, as $\{ \vh^1_v, \vh^2_v, \dots, \vh^T_v \}, \vh^t_v \in \sR^{f}$, which feed as input to the \emph{temporal} attention block.

\textbf{Temporal attention block.}
First, we equip the temporal attention module with a sense of ordering through \emph{position} embeddings~\citep{convs2s}, $ \{ \vp^1, \dots, \vp^T \}, \vp^t \in \sR^f$, which embed the absolute temporal position of each snapshot. The position embeddings are then combined with the output of the structural attention block to obtain a sequence of input representations: $\{ \vh^1_v + \vp^1, \vh^2_v + \vp^2, \dots,  \vh^T_v + \vp^T \}$ for node $v$ across multiple time steps.
This block also follows a similar structure with multiple stacked temporal self-attention layers. The outputs of the final layer pass into a position-wise \emph{feed-forward} layer to give the final node representations $\{ \ve^1_v, \ve^2_v, \dots,  \ve^T_v \} \; \forall v \in V$.

\textbf{Graph context prediction.}
To ensure that the learned representations capture both structural and temporal information, we define an objective function that preserves the local structure around a node, across multiple time steps. We use the dynamic representations of a node $v$ at time step $t$, $\ve_v^t$ to predict the occurrence of nodes appearing the local neighborhood around $v$ at $t$.
In particular, we use a binary cross-entropy loss function at each time step to encourage nodes co-occurring in fixed-length random walks, to have similar representations. 
\vspace{-2pt}
\begin{equation}
L_v = \sum\limits_{t=1}^T \sum\limits_{u \in \mathcal{N}^t_{walk} (v)}  - \log (\sigma ( <\ve^t_u, \ve^t_v>))-  w_n \cdot \sum\limits_{u^{'} \in P^t_n (v)} \log (1 - \sigma (<\ve^t_{u^{'}}, \ve^t_{v}>))
\label{eqn:loss}
\vspace{-5pt}
\end{equation}

where $\sigma$ is the sigmoid function, $<.>$ denotes the inner product operation, 
$\mathcal{N}^t_{walk} (v)$ is the set of nodes that co-occur with $v$ on fixed-length random walks at snapshot $t$, $P^t_n$ is a negative sampling distribution for snapshot $\gG^t$, and $w_n$, negative sampling ratio, is a tunable hyper-parameter to balance the positive and negative samples.

\section{Experiments}
We evaluate the quality of our learned node representations on the fundamental task of dynamic link prediction.
We choose this task since it has been widely used~\citep{know-evolve,dyngem,stream} in evaluating the quality of dynamic node representations to predict the temporal evolution in graph structure.

In our experiments, we compare the performance of~\name~against a variety of static and dynamic graph representation learning baselines.
Our experimental results on four publicly available benchmarks indicate that ~\name~ achieves significant performance gains over other methods.
\begin{table}[t] 
\centering
\begin{tabular}{@{}p{0.15\linewidth}K{0.15\linewidth}K{0.15\linewidth}K{0.15\linewidth}K{0.15\linewidth}K{0.15\linewidth}K{0.15\linewidth}@{}}
\toprule
\multirow{2}{*} & \multicolumn{2}{c}{\textbf{Communication Networks}}  &   \multicolumn{2}{c}{\textbf{Rating Networks} }\\
\cmidrule(lr){2-3} \cmidrule(lr){4-5}
\textbf{Dataset} & \textbf{Enron} & \textbf{UCI} & \textbf{Yelp} & \textbf{ML-10M} \\
\midrule
\textbf{\# Nodes} & 143 & 1,809 & 6,569 &  20,537\\
\textbf{\# Links} & 2,347 & 16,822 & 95,361 &  43,760 \\
\textbf{\# Time steps} & 10 & 13 & 12 & 13\\
 \bottomrule
\end{tabular}
\vspace{-5pt}
\caption{Statistics of the datasets used in our experiments}
\label{tab:dataset_stats}
\vspace{-10pt}
\end{table}
\subsection{Datasets}
We use four dynamic graph datasets with two communication and bipartite rating networks each. 

\textbf{Communication networks.} We consider two publicly available communication network datasets: Enron~\citep{enron} and UCI~\citep{UCI}.
In Enron, the communication links are email interactions between core employees and the links in UCI represent messages sent between users on an online social network platform. 

\textbf{Rating networks.}
We use two bipartite rating networks from Yelp\footnote{https://www.yelp.com/dataset/challenge} and MovieLens~\citep{ml-10m}.
In Yelp, the dynamic graph comprises links between two types of nodes, users and businesses, derived from the observed ratings over time.
ML-10M consists of a user-tag interaction network where user-tag links connects users with the tags they applied on certain movies.

In each dataset, multiple graph snapshots are created based on the observed interactions in fixed-length time windows.
Dataset statistics are shown in Table~\ref{tab:dataset_stats}, while Appendix~\ref{sec:dataset_details} has further details.

\subsection{Experimental Setup}
We conduct experiments on the task of link prediction in dynamic graphs, where we learn dynamic node representations on snapshots $\{\gG^1, \dots, \gG^t\}$ and use $\{ \ve_v^t, \forall v \in \gV\}$ to predict the links at $\gG^{t+1}$ during evaluation.
We compare different models based on their ability to correctly classify each example (node pair) into links and non-links.
\add{To further analyze predictive capability, we also evaluate \textit{new} link prediction, with a focus on new links that appear at each time step, (Appendix~\ref{sec:newlink}).}

We evaluate the performance of different models by training a logistic regression classifier for dynamic link prediction~\citep{dynamictriad}.
We create evaluation examples from the links in $\gG^{t+1}$ and an equal number of randomly sampled pairs of unconnected nodes (non-links).
A held-out validation set (20\% links) is used to tune the hyper-parameters across all models, which is later discarded.
We randomly sample 25\% of the examples for training and use the remaining 75\%  as our test set.
We repeat this for 10 randomized runs and report the average performance in our results. 

We follow the strategy recommended by ~\citet{node2vec} to compute the feature representation for a pair of nodes, using the Hadamard Operator ($\ve^t_u \odot \ve^t_v$), for all methods unless explicitly specified otherwise.
The Hadamard operator computes the element-wise product of two vectors and closely mirrors the widely used inner product operation in learning node embeddings. 
We evaluate the performance of link prediction using Area Under the ROC Curve (AUC) scores~\citep{node2vec}. 

 %

We implement~\name~in Tensorflow~\citep{tensorflow} and employ mini-batch gradient descent with Adam optimizer~\citep{adam} for training. 
For Enron, we use a single layer in both the structural and temporal blocks, with each layer comprising 16 attention heads computing 8 features apiece (for a total of 128 dimensions).
In the other datasets, we use two structural self-attentional layers with 16 and 8 heads respectively, each computing 16 features (layer sizes of 256, 128). 
The model is trained for a maximum of 200 epochs with a batch size of 256 nodes and the best performing model on the validation set, is chosen for evaluation.

\begin{table}[t] 
\centering
\scriptsize
\noindent\setlength\tabcolsep{2.3pt}
\begin{tabular}{@{}p{0.115\linewidth}K{0.099\linewidth}K{0.099\linewidth}K{0.099\linewidth}K{0.099\linewidth}K{0.105\linewidth}K{0.098\linewidth}K{0.098\linewidth}K{0.098\linewidth}@{}}
\toprule
\multirow{2}{*}{\textbf{Method}} & \multicolumn{2}{c}{\textbf{Enron}}  &   \multicolumn{2}{c}{\textbf{UCI} } & \multicolumn{2}{c}{\textbf{Yelp} }  & \multicolumn{2}{c}{\textbf{ML-10M} } \\
\cmidrule(lr){2-3} \cmidrule(lr){4-5} \cmidrule(lr){6-7} \cmidrule(lr){8-9}
 & \textbf{Micro-AUC} & \textbf{Macro-AUC} &  \textbf{Micro-AUC} & \textbf{Macro-AUC}  & \textbf{Micro-AUC} & \textbf{Macro-AUC} &  \textbf{Micro-AUC} & \textbf{Macro-AUC} \\

\midrule
node2vec & 83.72 $\pm\; 0.7$ & 83.05 $\pm \; 1.2$ & 79.99 $\pm \; 0.4$ & 80.49 $\pm\; 0.6$  & 67.86 $\pm \; 0.2$ & 65.34 $\pm \; 0.2$  & 87.74 $\pm \;0.2$ &  87.52 $\pm \;0.3$\\
G-SAGE & 82.48$^{*}\pm\; 0.6$ & 81.88$^{*} \pm 0.5$ & 79.15$^{*} \pm \; 0.4$ & 82.89$^{*} \pm \; 0.2$ & 60.95$^{\dagger}$ $\pm \; 0.1$  & 58.56$^{\dagger}\pm \; 0.2$  & 86.19$^{\ddagger}\pm \;0.3$ & 89.92$^{\ddagger}\pm \;0.1$ \\
G-SAGE + GAT & 72.52 $\pm\; 0.4$ & 73.34 $\pm \; 0.6$ &  74.03 $\pm \; 0.4$ & 79.83 $\pm\; 0.2$ & 66.15 $\pm \; 0.1$ & 65.09 $\pm \; 0.2$ & 83.97 $\pm \;0.3$ & 84.93 $\pm \;0.1$\\
\add{GCN-AE} & 81.55 $\pm\; 1.5$ & 81.71 $\pm \; 1.5$ & 80.53 $\pm \; 0.3$ & 83.50 $\pm\; 0.5$  & 66.71 $\pm \; 0.2$ & 65.82 $\pm \; 0.2$  & 85.49 $\pm \; 0.1$ &  85.74 $\pm \; 0.1$\\
\add{GAT-AE} & 75.71 $\pm\; 1.1$ & 75.97 $\pm \; 1.4$ & 79.98 $\pm \; 0.2$ & 81.86 $\pm\; 0.3$  & 65.92 $\pm \; 0.1$ & 65.37 $\pm \; 0.1$  & 87.01 $\pm \; 0.2$ &  86.75 $\pm \; 0.2$\\
\midrule
DynamicTriad & 80.26 $\pm\; 0.8$ & 78.98 $\pm \; 0.9$ & 77.59 $\pm \; 0.6$ & 80.28 $\pm\; 0.5$  & 63.53 $\pm \; 0.3$ & 62.69 $\pm \; 0.3$ & 88.71 $\pm \; 0.2$ & 88.43 $\pm \;0.1$\\
DynGEM & 67.83 $\pm \; 0.6$ & 69.72 $\pm \; 1.3$ & 77.49 $\pm \; 0.3$ & 79.82 $\pm\; 0.5$  & 66.02 $\pm \; 0.2$ & 65.94 $\pm \; 0.2$  & 73.69 $\pm \; 1.2$ & 85.96 $\pm \;0.3$\\
DynAERNN & 72.02 $\pm\; 0.7$ & 72.01 $\pm \; 0.7$ & 79.95 $\pm \; 0.4$ & 83.52 $\pm\; 0.4$  & 69.54 $\pm \; 0.2$ & 68.91 $\pm \; 0.2$ & 87.73 $\pm \;0.2$ & 89.47 $\pm \; 0.1$\\
\bf{DySAT} & \bf{85.71} $\pm \; 0.3$ & \bf{86.60} $\pm \; 0.2$ & \bf{81.03} $\pm\; 0.2$ & \bf{85.81} $\pm \; 0.1$  & \bf{70.15} $\pm \; 0.1$ & \bf{69.87} $\pm \; 0.1$ & \bf{90.82} $\pm \;0.3$ & \bf{93.68} $\pm \;0.1$ \\
 \bottomrule
\end{tabular}
\vspace{-5pt}
\caption{Experiment results on single-step link prediction (micro and macro averaged AUC with standard deviation). We show GraphSAGE (denoted by G-SAGE) results with the best performing aggregators for each dataset ($*$ represents GCN, $\dagger$ represents LSTM, and $\ddagger$ represents max-pooling).}
\label{tab:results}
\vspace{-10pt}
\end{table}

\subsection{Baseline}
We compare the performance of~\name~with several state-of-the-art dynamic graph embedding techniques.
In addition, we include several static graph embedding methods in comparison to analyze the gains of using temporal information for dynamic link prediction.
To make a fair comparison with static methods, we provide access to the entire history of snapshots by constructing an aggregated graph upto time $t$, where the weight of each link is defined as the cumulative weight till $t$ agnostic of its occurrence times.
We use author-provided implementations for all the baselines and set the final embedding dimension $d = 128$.

We compare against several state-of-the-art unsupervised static embedding methods: node2vec~\citep{node2vec}, GraphSAGE~\citep{graphsage} \add{and graph autoencoders~\citep{graph_enc_dec}}.
We experiment with different aggregators in GraphSAGE, namely, GCN, mean-pooling, max-pooling, and LSTM, to report the performance of the best performing aggregator in each dataset.
To provide a fair comparison with GAT~\citep{gat}, which originally conduct experiments only on node classification, we implement a graph attention layer as an additional aggregator in GraphSAGE, which we denote by GraphSAGE + GAT.
\add{We also train GCN and GAT as autoencoders for link prediction along the suggested lines of~\citep{gcn_lp}, denoted by GCN-AE and GAT-AE respectively.}
In the dynamic setting, we evaluate~\name~ against the most recent studies on dynamic graph embedding including DynAERNN~\citep{dyngraph2vec}, DynamicTriad~\citep{dynamictriad}, and DynGEM~\citep{dyngem}.
The details of hyper-parameter tuning for all methods can be found in Appendix~\ref{sec:hyper_param_details}.

\subsection{Experimental Results}
We evaluate the models at each time step $t$ by training separate models up to snapshot $t$ and evaluate at $t+1$ for each $t = 1, \dots, T$.
We summarize the micro and macro averaged AUC scores (across all time steps) for all models in Table~\ref{tab:results}.
From the results, we observe that ~\name~ achieves consistent gains of 3--4\% macro-AUC, in comparison to the best baseline across all datasets.

\begin{figure}[tb]
\centering
\includegraphics[width=1.0\linewidth]{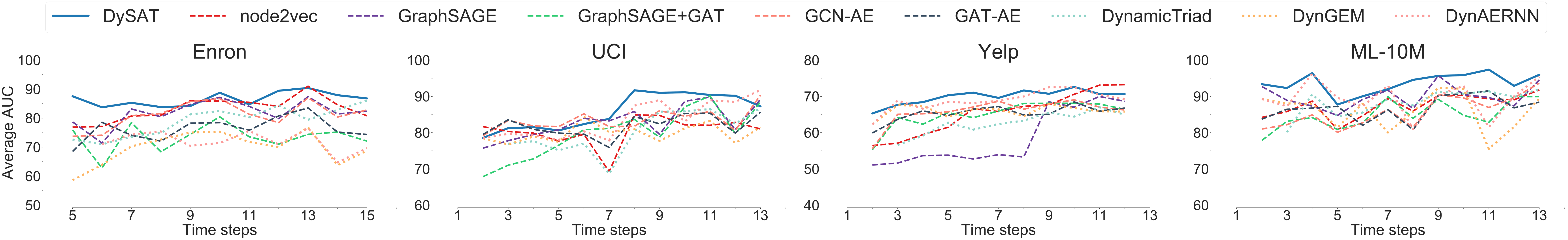}
\vspace{-20pt}
\caption{Performance comparison of~\name~with different models across multiple time steps: the solid line represents~\name; dashed lines represent static graph embedding models; and dotted lines represent dynamic graph embedding models. We truncate the y-axis to avoid visual clutter.}
\label{fig:results}
\vspace{-15pt}
\end{figure}

Further, we compare the model performance at each time step (Figure~\ref{fig:results}), to obtain a deep understanding of their temporal behaviors. 
We fine the performance of~\name~to be relatively more stable than other methods.
This contrast is pronounced in the communication networks (Enron and UCI), where we observe drastic drops in performance of static embedding methods at certain time steps.

\add{The runtime per mini-batch of~\name~on ML-10M, using a machine with Nvidia Tesla V100 GPU and 28 CPU cores, is 0.72 seconds. In comparison, a model variant without temporal attention (Appendix~\ref{sec:temporal_analysis}) takes 0.51 seconds, which illustrates the relatively low cost of temporal attention.}

\section{Discussion}
Our experimental results provide several interesting observations and insights to the performance of different graph embedding techniques.

First, we observe that GraphSAGE achieves comparable performance to DynamicTriad across different datasets, despite being trained only on static graphs.
One possible explanation may be that GraphSAGE uses trainable neighbor-aggregation functions, while DynamicTriad employs Skip-gram based methods augmented with temporal smoothness.
This leads us to conjecture that the combination of structural and temporal modeling with expressive aggregation functions, such as multi-head attention, is responsible for the consistently superior performance of~\name~on dynamic link prediction.
We also observe that node2vec achieves consistent performance agnostic of temporal information, which demonstrates the effectiveness of second-order random walk sampling.
This observation points to the direction of applying sampling techniques to further improve~\name.

In~\name, we employ structural attention layers followed by temporal attention layers.
We choose this design because graph structures are not stable over time, which makes directly employing structural attention layers after temporal attention layers infeasible.
We also consider another alternative design choice that applies self-attention along the two dimensions of neighbors and time together following the strategy similar to~\citep{disan}.
In practice, this would be computationally expensive due to variable number of neighbors per node across multiple snapshots. 
We leave exploring other architectural design choices based on structural and temporal self-attentions as future work. 

In the current setup, we store the adjacency matrix of each snapshot in memory using sparse matrix, which may pose memory challenges when scaling to large graphs.
In the future, we plan to explore~\name~with memory-efficient mini-batch training strategy along the lines of GraphSAGE~\citep{graphsage}. 
Further, we develop an incremental self-attention network (IncSAT) that is efficient in both computation and memory cost as a direct extension of~\name.
Our initial results are promising as reported in Appendix~\ref{sec:incsat}, which opens the door to future exploration of self-attentional architectures for incremental (or streaming) graph representation learning.
\add{We also evaluate the capability of~\name~on multi-step link prediction or forecasting and observe significant relative improvements of 6\% AUC on average over existing methods, as reported in Appendix~\ref{sec:forecast}.}

\section{Conclusion}
In this paper, we introduce a novel self-attentional neural network architecture named~\name~to learn node representations in dynamic graphs.
Specifically, ~\name~computes dynamic node representations using self-attention over the (1) structural neighborhood and (2) historical node representations, thus effectively captures the temporal evolutionary patterns of graph structures.
Our experiment results on various real-world dynamic graph datasets indicate that~\name~achieves significant performance gains over several state-of-the-art static and dynamic graph embedding baselines.
Though our experiments are conducted on graphs without node features,~\name~can be easily generalized on feature-rich graphs.
Another interesting direction is exploring continuous-time generalization of our framework to incorporate more fine-grained temporal variations.

\bibliography{main-arxiv}
\bibliographystyle{iclr2019_conference}
\appendix
\newpage

\section{Self-Attention Analysis}
Since self-attention plays a critical role in the formulation of~\name, we conduct exploratory analyses to study 
(a) \textit{effectiveness}: how do the different attention layers affect the performance of~\name? 
(b) \textit{efficiency}: how efficient is  self-attention compare in comparison to existing recurrent models? and 
(c) \textit{interpretability}: are the learnt attention weights visually interpretable for humans?
\subsection{Effectiveness of Self-Attention}
\label{sec:temporal_analysis}
\begin{figure}[ht]
\includegraphics[width=\linewidth]{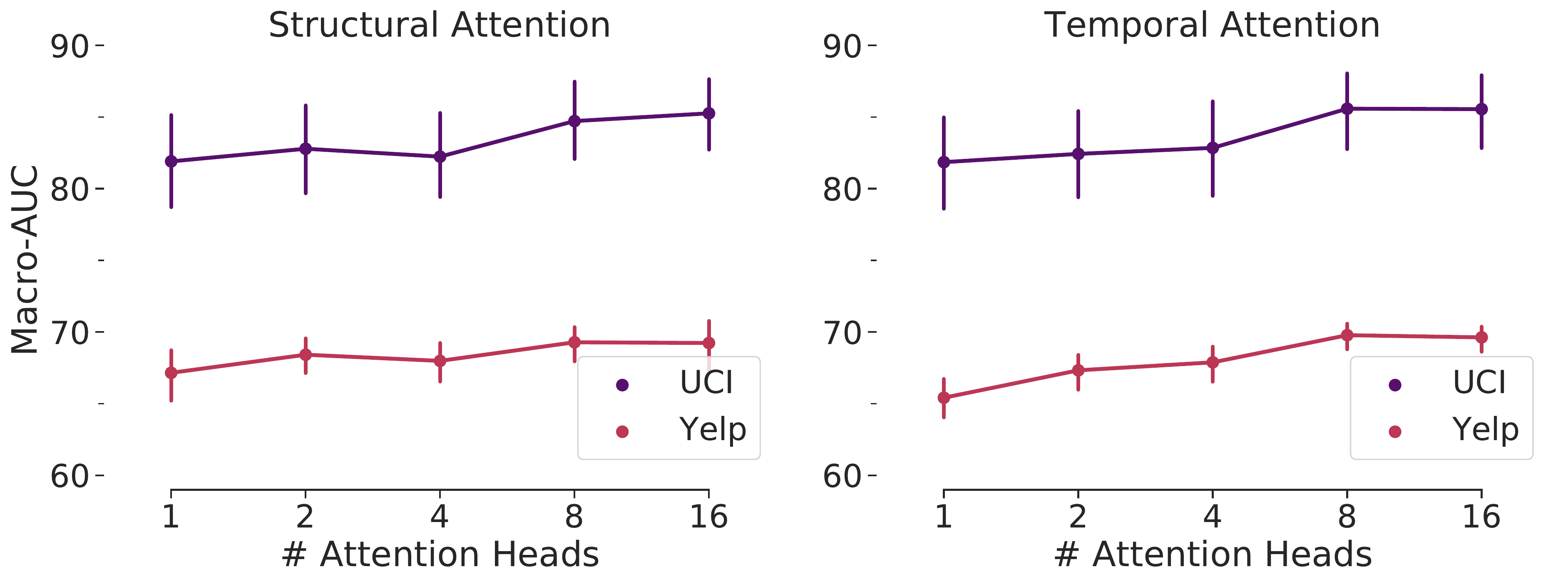}
\vspace{-15pt}
\caption{Sensitivity analysis on the number of structural and temporal attention heads in~\name~(UCI and Yelp)}
\label{fig:heads}
\vspace{-10pt}
\end{figure}

We evaluate the effectiveness of our self-attentional architecture~\name~in two parts:\\
\begin{itemize}[leftmargin=*]
\item \textbf{Structural and temporal self-attention effectiveness.}
We present an ablation study on the structural and temporal attentional layers in~\name. 
To demonstrate the effectiveness of self-attentional layers, we independently remove the structural and temporal attention blocks from~\name~to create simpler architectures, that are optimized using the same loss function (Eqn.~\ref{eqn:loss}).
Note that the model obtained on removing the temporal attention block is different from static methods since the node embeddings are jointly optimized (using Eqn.~\ref{eqn:loss}) across all snapshots without any explicit temporal modeling.
We use the same hyper-parameter configuration of~\name~on each dataset from the original experiments to train the new model.
The performance comparison is shown in Table~\ref{tab:notemp}.

We observe that in some datasets, the structural attention block is able to learn some temporal evolution patterns in graph structure, despite the lack of explicit temporal modeling, while the removal of structural attention leads to a decrease in model performance on most datasets.
However, DySAT consistently outperforms the variants with  3\% average gain in Macro-AUC, which validates our choice of using structural and temporal self-attentional layers.\\

\item \textbf{Multi-head attention effectiveness.}
We conduct a sensitivity analysis on the number of attention heads used during multi-head attention.
Specifically, we vary the number of structural and temporal heads in~\name~independently in the range $\{ 1, 2, 4, 8, 16\}$, while keeping the layer sizes fixed for fairness. 
As shown in Figure~\ref{fig:heads}, we can see that~\name~benefits from multi-head attention on both structural and temporal layers.
The performance stabilizes with 8 attention heads, which appears sufficient to capture graph evolution from multiple latent facets.
\end{itemize}
 
\begin{table}[h] 
\centering
\scriptsize
\noindent\setlength\tabcolsep{2.5pt}
\begin{tabular}{@{}p{0.1\linewidth}K{0.098\linewidth}K{0.098\linewidth}K{0.098\linewidth}K{0.098\linewidth}K{0.105\linewidth}K{0.098\linewidth}K{0.098\linewidth}K{0.098\linewidth}@{}}
\toprule
\multirow{2}{*}{\textbf{Method}} & \multicolumn{2}{c}{\textbf{Enron}}  &   \multicolumn{2}{c}{\textbf{UCI} } & \multicolumn{2}{c}{\textbf{Yelp} }  & \multicolumn{2}{c}{\textbf{ML-10M} } \\
\cmidrule(lr){2-3} \cmidrule(lr){4-5} \cmidrule(lr){6-7} \cmidrule(lr){8-9}
 & \textbf{Micro-AUC} & \textbf{Macro-AUC} &  \textbf{Micro-AUC} & \textbf{Macro-AUC}  & \textbf{Micro-AUC} & \textbf{Macro-AUC} &  \textbf{Micro-AUC} & \textbf{Macro-AUC} \\

\midrule
Original & \bf{85.71} $\pm\; 0.3$ & \bf{86.60} $\pm \; 0.2$ & \bf{81.03} $\pm\; 0.2$ & \bf{85.81} $\pm \; 0.1$  & \bf{70.15} $\pm \; 0.1$ & \bf{69.87} $\pm \; 0.1$ & \bf{90.82}  $\pm \;0.3$ & \bf{93.68} $\pm \;0.1$ \\

No Structural & 85.21 $\pm\; 0.2$  & 86.50 $\pm\; 0.3$ & 74.48  $\pm \; 0.3$ & 81.24 $\pm \; 0.2$ & 70.11 $\pm \; 0.1$ & 67.85  $\pm \; 0.1$ &  88.56 $\pm \;0.2$ & 90.34 $\pm \;0.2$\\

No Temporal & 84.50 $\pm\; 0.3$  & 85.68 $\pm\; 0.4$ & 76.61 $\pm \; 0.2$ & 79.97 $\pm\; 0.3$ & 68.34  $\pm \; 0.1$ & 67.20 $\pm \; 0.3$ & 89.61 $\pm \; 0.4$  &  91.10  $\pm \; 0.2$ \\

\bottomrule
\end{tabular}
\vspace{-5pt}
\caption{Ablation study on structural and temporal attention layers (micro and macro averaged AUC with std. deviation)}
\vspace{-5pt}
\label{tab:notemp}
\end{table}

\subsection{Scalability Analysis}
\label{sec:runtime}
\begin{figure}[h]
\centering
\vspace{-5pt}
\subfigure[]{
\includegraphics[width=0.35\linewidth]{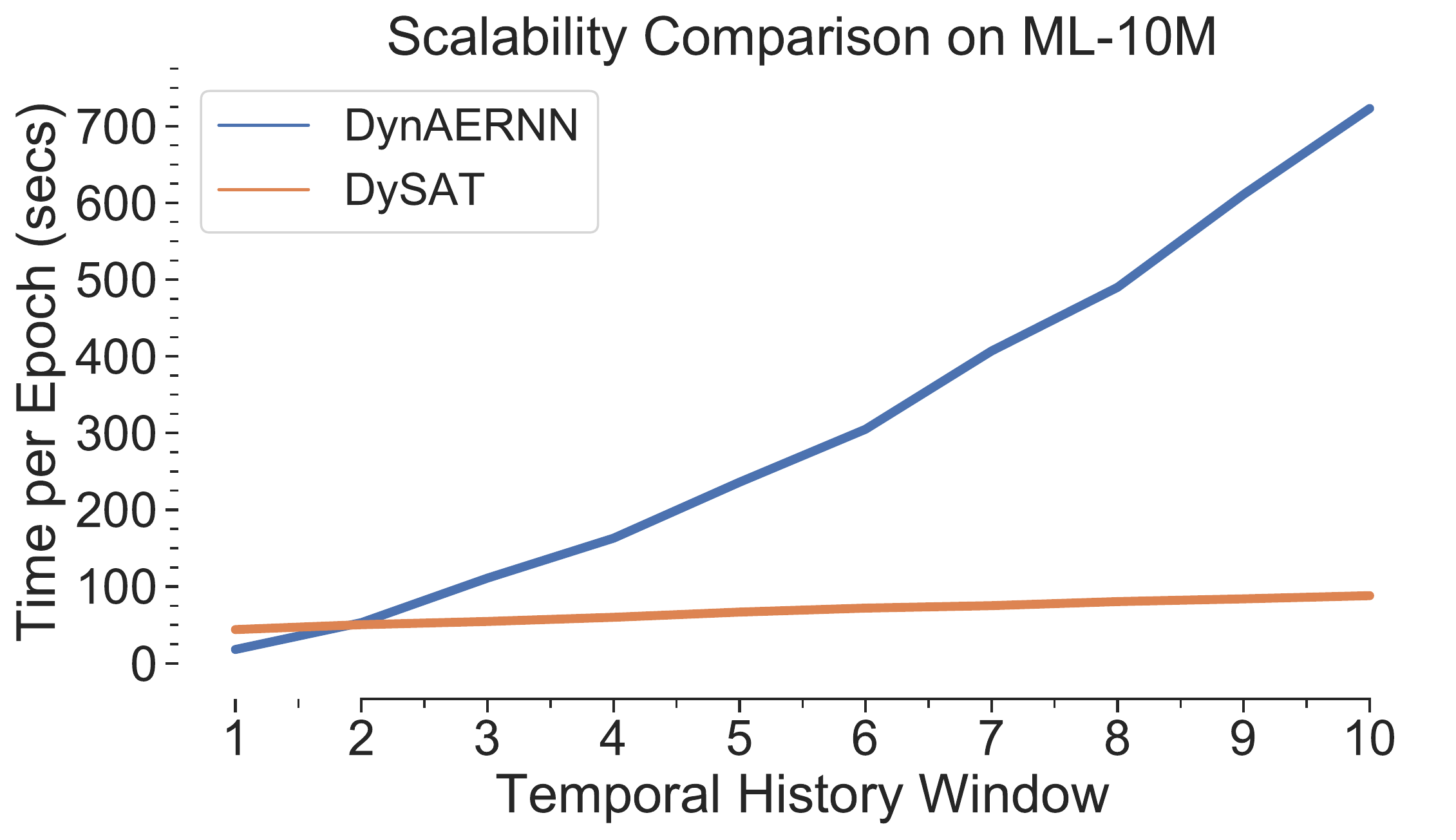}
\vspace{-10pt}
\label{fig:runtime}}
\subfigure[]{
\includegraphics[width=0.55\linewidth]{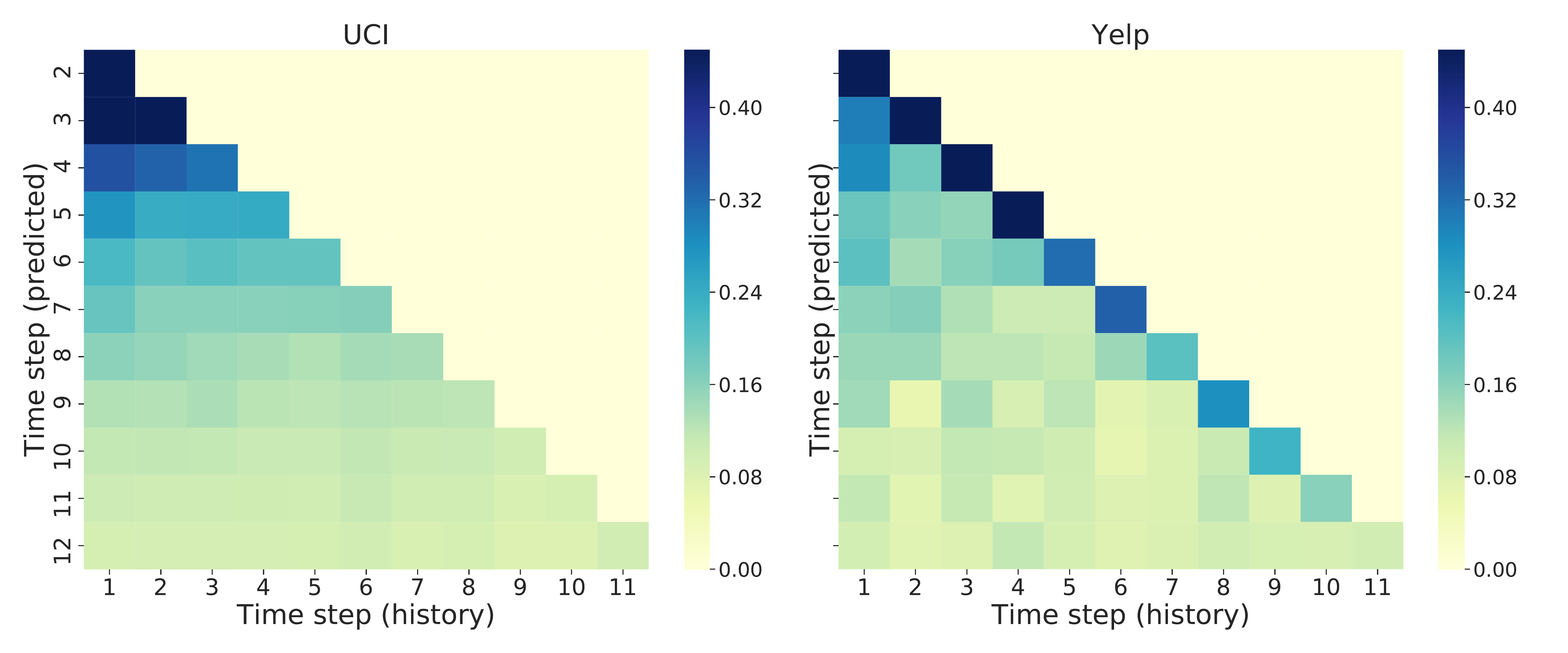}
\vspace{-15pt}
\label{fig:attention}
}
\vspace{-5pt}
\caption{(a) Scalability comparison of~\name~with DynAERNN. (b) Heatmap of the mean temporal attention weights on UCI and Yelp dataset.}
\vspace{-15pt}
\end{figure}

In this section, we compare the scalability of~\name~against a state-of-the-art dynamic embedding method DynAERNN, which uses an RNN-based architecture to model dynamic graph evolution.
We choose this method for comparison as it is conceptually closest to our self-attentional architecture, as per well-established literature.
We compare the model training time of two methods by varying the temporal history \textit{window}, \textit{i.e.}, the number of previous snapshots used as temporal history.
Figure~\ref{fig:runtime} depicts runtime per epoch of both models on ML-10M, using a machine with Nvidia Tesla V100 GPU and 16 CPU cores.

~\name~achieves significantly lower training times than DynAERNN, \textit{e.g.},
the runtime per epoch of~\name~is 88 seconds with a window size of 10 on ML-10M, in comparison to 723 seconds for DynAERNN.
Self-attention is easily parallelized across both time as well as attention heads, while RNNs suffer due to the inherently sequential nature of back-propagation through time.
Our results demonstrate the practical scalability advantage for pure self-attentional architectures over RNN-based methods.

\subsection{Temporal Attention Visualization}
We also conduct a qualitative analysis attempting to obtain deeper insights into the distribution of temporal attention weights learned by~\name~on datasets with distinct evolutionary behaviors.
In this experiment, we examine the temporal attention coefficients learned at each time step $k$, which indicate the relative importance of each historical snapshot in predicting the links at $k$.
We consider two diverse datasets UCI and Yelp: UCI is an email communication network with periodically recurring user interactions, while Yelp is a rating network where new user-business ratings get added over time.
Figure~\ref{fig:attention} visualizes a heatmap of the learned temporal attention weights on the UCI and Yelp dataset for the first 11 time steps.

In Figure~\ref{fig:attention}, each row depicts the set of all attention weights over historical time steps $t_1 ... t_{k-1}$ for predicting links at time step $k$.
We find the attention weights to be biased towards recent snapshots in Yelp, while being more uniformly distributed in UCI.
This observation conforms to the nature of graph evolution in these datasets since rating behaviors in Yelp tends to be bursty and correlated with events such restaurant opening, discounted sales, etc., while inter-user communications in UCI typically span longer time intervals.
Thus,~\name~is able to learn different distributions of attention weights 
adaptive to the mechanism of temporal graph evolution across different datasets.
While this analysis provides a macro-perspective on the weights learned by temporal attention across different datasets,
an appropriate node-level interpretation of these coefficients (such as \textit{e.g.},~\citep{bahdanau}) requires further domain knowledge about the dataset under study, and is left as future work.

We conduct a qualitative analysis to obtain deeper insights into the distribution of temporal attention weights learned by~\name.
In this experiment, we examine the temporal attention coefficients learned at each time step $t$, which indicate the relative importance of each historical snapshot ($<t$) in predicting the links at $t$.
We choose the Enron dataset to visualize the mean and standard deviation of temporal attention coefficients, over all the nodes.
Figure~\ref{fig:attention} visualizes a heatmap of the learned temporal attention weights on Enron dataset for the first 10 time steps.

From Figure~\ref{fig:attention}, we observe that the mean temporal attention weights are mildly biased towards recent snapshots, while the historical snaphots vary in their importance across different time steps.
Further, we find that the standard deviation of attention weights across different nodes is generally high and exhibits more variability. 
Thus, the temporal attention weights are well distributed across historical snapshots, with significant variance across different nodes in the graph.
While this analysis attempts to provide a high-level perspective on the weights learned by temporal attention,
an appropriate interpretation of these coefficients (as done by \textit{e.g.},~\citep{bahdanau}) requires further domain knowlege about the dataset under study, and is left as future work.


\section{Dynamic New Link Prediction}
\label{sec:newlink}
In this section, we additionally report the results of dynamic link prediction evaluated only on the \textit{new} links at each time step. This provides an in-depth analysis on the capabilities of different methods in predicting relatively unseen links. We follow the same evaluation setup of training a downstream logistic regression classifier for dynamic link prediction.
However, a key difference is that the evaluation examples comprise new links at $\gG_{t+1}$ (that are not in $\gG_t$) and an equal number of randomly sampled non-links.

Table~\ref{tab:newlink_results} summarizes the micro and macro averaged AUC scores for different methods on the four datasets.
The absolute performance numbers of all methods are lower than the original evaluation setup of using all links at $\gG_{t+1}$, which is reasonable since accurate prediction of new links at $\gG_{t+1}$  is expected to be slightly more challenging in comparison to predicting all the links at $\gG_{t+1}$.
From Table~\ref{tab:results}), we find that~\name~achieves consistent relative gains of 3--5\% Macro-AUC over the best baselines on dynamic new link prediction as well, thus validating its effectiveness in accurately capturing temporal context for new link prediction.
\begin{table}[h] 
\centering
\scriptsize
\noindent\setlength\tabcolsep{2.3pt}
\begin{tabular}{@{}p{0.115\linewidth}K{0.099\linewidth}K{0.099\linewidth}K{0.099\linewidth}K{0.099\linewidth}K{0.105\linewidth}K{0.098\linewidth}K{0.098\linewidth}K{0.098\linewidth}@{}}
\toprule
\multirow{2}{*}{\textbf{Method}} & \multicolumn{2}{c}{\textbf{Enron}}  &   \multicolumn{2}{c}{\textbf{UCI} } & \multicolumn{2}{c}{\textbf{Yelp} }  & \multicolumn{2}{c}{\textbf{ML-10M} } \\
\cmidrule(lr){2-3} \cmidrule(lr){4-5} \cmidrule(lr){6-7} \cmidrule(lr){8-9}
 & \textbf{Micro-AUC} & \textbf{Macro-AUC} &  \textbf{Micro-AUC} & \textbf{Macro-AUC}  & \textbf{Micro-AUC} & \textbf{Macro-AUC} &  \textbf{Micro-AUC} & \textbf{Macro-AUC} \\

\midrule
\add{node2vec} & 76.92 $\pm\; 1.2$ & 75.86 $\pm \; 0.5$ & 73.67 $\pm \; 0.3$ & 74.76 $\pm\; 0.8$  & 67.36 $\pm \; 0.2$ & 65.17 $\pm \; 0.2$  & 85.22 $\pm \;0.2$ &  84.89 $\pm \;0.1$\\
\add{G-SAGE} & 73.92$^{*}\pm\; 0.7$ & 74.67$^{*} \pm 0.6$ & 76.69$^{*} \pm \; 0.3$ & 79.41$^{*} \pm \; 0.1$ & 62.25$^{\dagger}$ $\pm \; 0.2$  & 58.81$^{\dagger}\pm \; 0.3$  & 85.23$^{\ddagger}\pm \;0.3$ & 89.14$^{\ddagger}\pm \;0.2$ \\
\add{G-SAGE + GAT} & 67.02 $\pm\; 0.8$ & 68.32 $\pm \; 0.7$ &  73.18 $\pm \; 0.4$ & 76.79 $\pm\; 0.2$ & 66.53 $\pm \; 0.2$ & 65.45 $\pm \; 0.1$ & 80.84 $\pm \;0.3$ & 82.53 $\pm \;0.1$\\
\add{GCN-AE} & 74.46 $\pm\; 1.1$ & 74.02 $\pm \; 1.6$ & 74.76 $\pm \; 0.1$ & 76.75 $\pm\; 0.6$  & 66.18 $\pm \; 0.2$ & 65.77 $\pm \; 0.3$  & 82.45 $\pm \; 0.3$ &  82.48 $\pm \; 0.2$\\
\add{GAT-AE} & 69.75 $\pm\; 2.2$ & 69.25 $\pm \; 1.9$ & 72.52 $\pm \; 0.4$ & 73.78 $\pm\; 0.7$  & 66.07 $\pm \; 0.1$ & 65.91 $\pm \; 0.2$  & 84.98 $\pm \; 0.2$ &  84.51 $\pm \; 0.3$\\
\midrule
\add{DynamicTriad} & 69.59 $\pm\; 1.2$ & 68.77 $\pm \; 1.7$ & 67.97 $\pm \; 0.7$ & 71.67 $\pm\; 0.9$ & 63.76 $\pm \; 0.2$ & 62.83 $\pm \; 0.3$ & 84.72 $\pm \; 0.2$ & 84.32 $\pm \;0.2$\\
\add{DynGEM} & 60.73 $\pm \; 1.1$ & 62.85 $\pm \; 1.9$ & 77.49 $\pm \; 0.3$ & 79.82 $\pm\; 0.5$  & 66.42 $\pm \; 0.2$ & 66.84 $\pm \; 0.2$  & 73.77 $\pm \; 0.7$ & 83.51 $\pm \;0.3$\\
\add{DynAERNN} & 59.05 $\pm\; 2.7$ & 59.63 $\pm \; 2.7$ & 77.72 $\pm \; 0.5$ & 81.91 $\pm\; 0.6$  & \bf{74.33} $\pm \; 0.2$ & \bf{73.46} $\pm \; 0.2$ & 87.42  $\pm \;0.2$ & 88.19 $\pm \; 0.2$\\
\add{\bf{DySAT}} & \bf{78.87} $\pm \; 0.6$ & \bf{78.58} $\pm \; 0.6$ & \bf{79.24} $\pm\; 0.3$ & \bf{83.66} $\pm \; 0.2$  & \bf{69.46} $\pm \; 0.1$ & \bf{69.14} $\pm \; 0.1$ & \bf{89.29} $\pm \;0.2$ & \bf{92.65} $\pm \;0.1$ \\
 \bottomrule
\end{tabular}
\vspace{-10pt}
\caption{\add{Experiment results on dynamic \textit{new} link prediction (micro and macro averaged AUC with standard deviation). We show GraphSAGE (denoted by G-SAGE) results with the best performing aggregators for each dataset ($*$ represents GCN, $\dagger$ represents LSTM, and $\ddagger$ represents max-pooling).}}
\vspace{-10pt}
\label{tab:newlink_results}
\end{table}

\add{
\section{Multi-Step Link Prediction}
\label{sec:forecast}
In this section, we evaluate various dynamic graph representation learning methods on the task of multi-step link prediction or forecasting. 
Here, each model is trained for a fixed number of time steps, and the latest embeddings are used to predict links at multiple future time steps.
In each dataset, we choose the last 6 snapshots to evaluate multi-step link prediction.
The model is trained on the previous remaining snapshots, and the latest embeddings are used to forecast links at future time steps.
For each future time step $t+\Delta \; (1 \leq  \Delta \leq 6)$, we create examples from the links in $\gG_{t+\Delta}$ and an equal number of randomly sampled pairs of unconnected nodes (non-links).
We otherwise use the same evaluation setup of training a downstream logistic regression classifier to evaluate link prediction.
In this experiment, we exclude the links formed by nodes that newly appear in the future evaluation snapshots, since most methods cannot be easily support updates for new nodes.
}

\begin{figure}[t]
\centering
\includegraphics[width=1.0\linewidth]{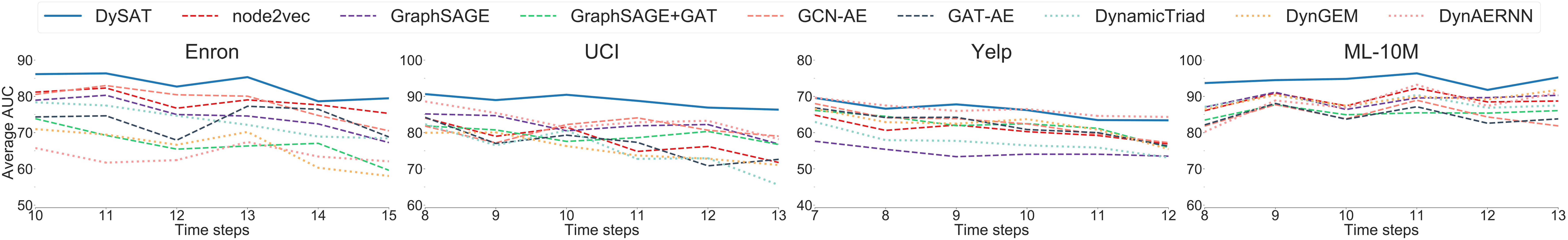}
\vspace{-20pt}
\caption{\add{Performance comparison of~\name~with different models on multi-step link prediction for 6 future time steps on all datasets}}
\label{fig:forecast}
\end{figure}

\add{Figure.~\ref{fig:forecast} depicts the variation in model performance of different methods over the 6 evaluation snapshots. 
As expected, we observe a slight decay in performance over time for all the models.
~\name achieves significant performance gains over all other baselines and maintains a highly stable link prediction performance over multiple future time steps.
Static embedding methods often exhibit large variations in performance over time steps, while~\name~ achieves a stable consistent performance.
The historical context captured by the dynamic node embeddings of~\name,~is one of the most likely reasons for its stable multi-step forecasting performance.}

\add{
\section{Impact of unseen nodes on Dynamic Link Prediction}
In this section, we analyze the sensitivity of different graph representation learning on link prediction for previously unseen nodes that appear newly at time $t$.}

\begin{figure}[h]
\centering
\includegraphics[width=1.0\linewidth]{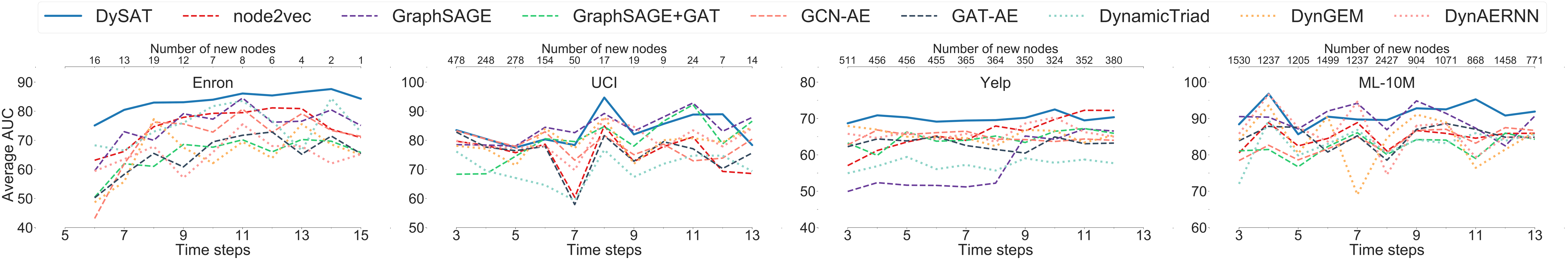}
\vspace{-10pt}
\caption{\add{Performance comparison of~\name~with different models on link prediction restricted to \textit{new} nodes at each time step}}
\label{fig:newnodes}
\end{figure}

\add{A node is considered as a \textit{new} node at time step $t$ in $\gG_t$ if it has not appeared (has no links) in any of the previous $t-1$ snapshots.
In this experiment, the evaluation set at time step $t$ only comprises the subset of links at $\gG_{t+1}$ among the new nodes in $\gG_t$ and corresponding randomly sampled non-links.
Since the number of nodes varies significantly across different time steps, we report the performance of each method along with the number of new nodes at each time step, in Figure~\ref{fig:newnodes}.}

\add{From Figure~\ref{fig:newnodes}, we observe that~\name~outperforms other baselines in most datasets, demonstrating the ability to characterize new or previously unseen nodes despite their limited history.
Although the temporal attention will focus on the latest representation of a new node $v$ due to absence of history, the structural embedding of $v$ recieves backpropagation signals through the temporal attention on neighboring nodes, which indirectly affects the final embedding of $v$.
We hypothesize that this indirect temporal signal is one of the reasons for~\name~to achieve performance improvements over baselines, albeit not designed to explicitly model historical context for previously unseen nodes.}

\begin{figure}[t]
\includegraphics[width=\linewidth]{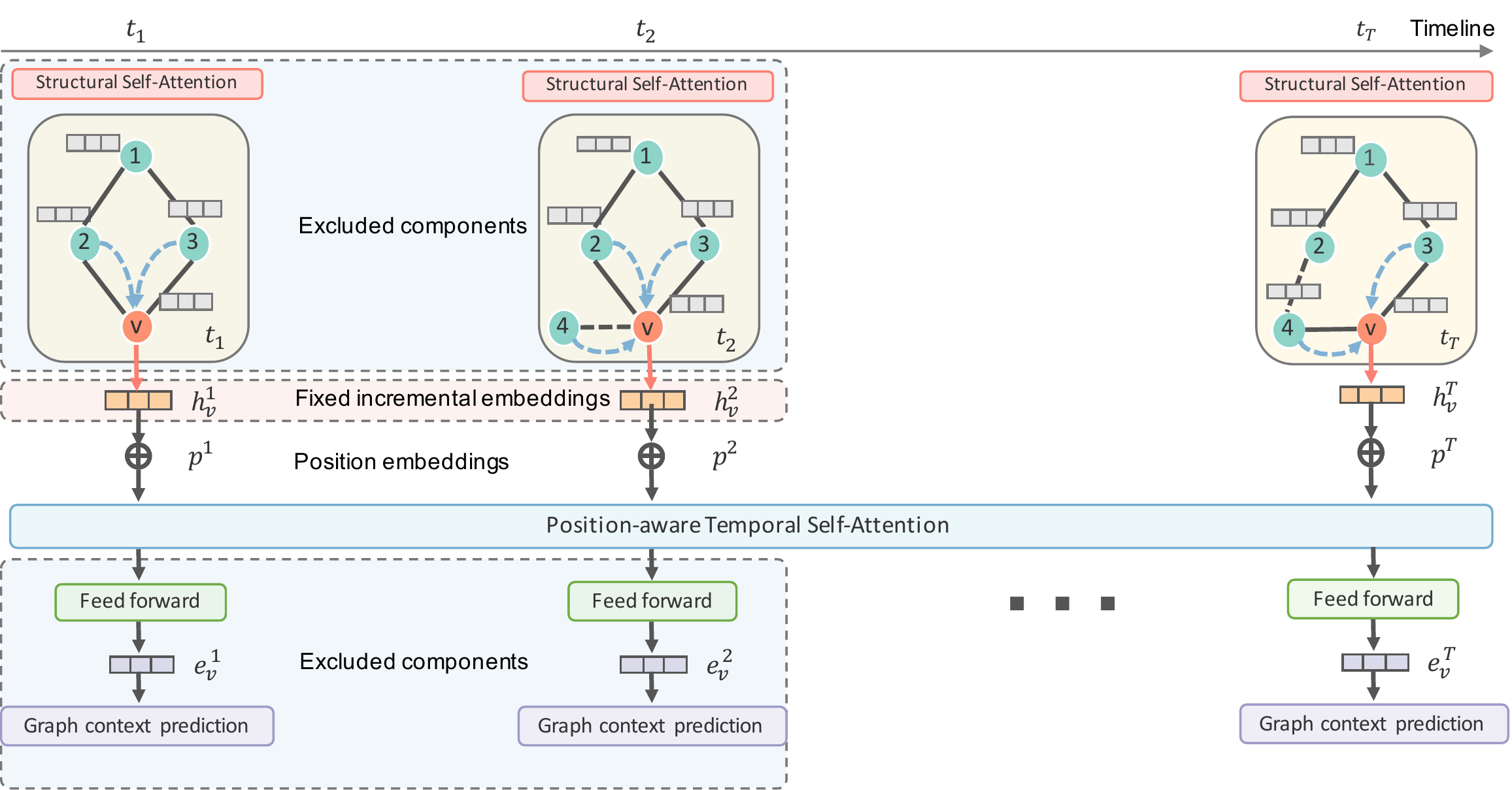}
\vspace{-15pt}
\caption{Neural architecture of IncSAT: the components that are excluded from~\name~are wrapped by dashed blue rectangles. The intermediate node representations are directly loaded from saved models trained previously.}
\label{fig:incsat}
\vspace{-10pt}
\end{figure}
\section{Incremental Self-Attention Network}
\label{sec:incsat}
In this section, we describe an extension of our dynamic self-attentional architecture to learn incremental node representations.
The motivation of incremental learning arises due to the proliferation in sizes of real-world graphs, making it difficult to store multiple snapshots in memory.
Thus, the incremental graph representation learning problem imposes the restriction of no access to historical graph snapshots, in contrast to most dynamic graph embedding methods.
Specifically, to learn the node embeddings $\{ \ve_v^t \in \sR^d \; \forall v \in \gV \}$ at time $T$, we require a model to only access to the snapshot $\gG^T$ and a \emph{summary} of the historical snapshots.
For example, DynGEM~\citep{dyngem} is an example of an incremental embedding method that uses the embeddings learned at step $t-1$, as initialization to learn the embeddings at $t$.

We propose an extension of our self-attentional architecture named IncSAT to explore solving the incremental graph representation learning problem.
To learn node representations at $T$, we first incrementally train multiple models at $1 \leq t \leq T$.
Unlike the original~\name~where structural self-attention is applied at each snapshot, IncSAT applies the structural block only at the latest graph snapshot $\gG^t$.
We enable incremental learning by storing the intermediate output representations $\{\vh^T_v \; \forall v \in \gV\}$ of the structural block.
As illustrated in Figure~\ref{fig:incsat}, these intermediate output representations of historical snapshots $(1 \leq t < T)$ can be directly loaded from previously saved results at $1 \leq t < T$. 
Thus, the structural information of the previous historical snapshots are \emph{summarized} in the stored intermediate representations.
The temporal self-attention is only applied to the current snapshot $\gG^T$ over the historical representations of each node to compute the final node embeddings $\{ \ve^T_v \; \forall v \in \gV\}$ at $T$, which are trained on random walks sampled from $\gG^T$.

We evaluate IncSAT using the same experimental setup, with minor modifications in hyper-parameters.
We use a dropout rate of 0.4 in both the structural and temporal self-attention layers.
From our preliminary experiments, we find that a higher dropout rate in the structural block can facilitate avoiding over-fitting the model to the current graph snapshot.
In Table~\ref{tab:incsat}, we report the performance of IncSAT in comparison to DySAT and DynGEM, which is the only one that can support incremental training from our baseline models. 
The results show that IncSAT achieves comparable performance to DySAT on most datasets while significantly outperforming DynGEM, albeit with minimal hyper-parameter tuning.

\begin{table}[t] 
\centering
\scriptsize
\noindent\setlength\tabcolsep{2.5pt}
\begin{tabular}{@{}p{0.1\linewidth}K{0.098\linewidth}K{0.098\linewidth}K{0.098\linewidth}K{0.098\linewidth}K{0.105\linewidth}K{0.098\linewidth}K{0.098\linewidth}K{0.098\linewidth}@{}}
\toprule
\multirow{2}{*}{\textbf{Method}} & \multicolumn{2}{c}{\textbf{Enron}}  &   \multicolumn{2}{c}{\textbf{UCI} } & \multicolumn{2}{c}{\textbf{Yelp} }  & \multicolumn{2}{c}{\textbf{ML-10M} } \\
\cmidrule(lr){2-3} \cmidrule(lr){4-5} \cmidrule(lr){6-7} \cmidrule(lr){8-9}
 & \textbf{Micro-AUC} & \textbf{Macro-AUC} &  \textbf{Micro-AUC} & \textbf{Macro-AUC}  & \textbf{Micro-AUC} & \textbf{Macro-AUC} &  \textbf{Micro-AUC} & \textbf{Macro-AUC} \\

\midrule
\bf{DySAT} & \bf{85.71} $\pm\; 0.3$ & \bf{86.60} $\pm \; 0.2$ & \bf{81.03} $\pm\; 0.2$ & \bf{85.81} $\pm\; 0.1$  & \bf{70.15} $\pm \; 0.1$ & \bf{69.87} $\pm \; 0.1$ & \bf{90.82} $\pm \; 0.3$ & \bf{93.68} $\pm \; 0.1$ \\
DynGEM & 67.83 $\pm\; 0.6$ & 69.72 $\pm \; 1.3$ & 77.49 $\pm \; 0.3$& 79.82 $\pm\; 0.5$  & 66.02 $\pm \; 0.2$ & 65.94 $\pm \; 0.2$  & 73.69 $\pm \;1.2$ & 85.96 $\pm \;0.3$\\
IncSAT & 84.36 $\pm \; 0.2$ & 85.43 $\pm \; 0.3$ & 76.18 $\pm \; 0.5$ & 85.37 $\pm \; 0.2$ & 69.54 $\pm \; 0.1$ & 68.73 $\pm \; 0.3$  & 80.13 $\pm \; 0.4$ &  91.14 $\pm \; 0.2$\\
 \bottomrule
\end{tabular}
\caption{Experimental results of IncSAT in comparison to~\name~(micro and macro averaged AUC with standard deviation)}
\label{tab:incsat}
\end{table}

\section{Computational Complexity}
The cost is split into two parts: (a) structural self-attention, which takes $O(|\gV| D^2 + |\gE_t| D)$ per snapshot, giving a total cost of $O(|\gV| T D^2 + \sum\limits_{t=1}^T |\gE_t| D )$ over $T$ snapshots,
(b) temporal self-attention, which includes the self-attentional and feed-forward layers to give $O(|\gV| T^2 D + |\gV| T D^2)$.
The overall complexity of~\name~is $O(|\gV| T^2 D  + |\gV| T D^2 + \sum\limits_{t=1}^T |\gE_t| D )$. The dominant term is $O(|\gV| T^2 D)$ from the temporal self-attention layer, however its computation is fully parallelizable, thus amenable to GPU acceleration.
In contrast, RNN-based methods (e.g., DynAERNN~\cite{dyngraph2vec}) have a sequential dependency on previous steps, \textit{i.e.}, computation at $t$ must wait from embeddings at $t-1$, resulting in $O(T)$ cost of sequential sequential operations.
Empirically, we find~\name~to be around ten times faster than RNN-based methods with GPUs.

\section{Details on Hyper-parameter Settings and Tuning}
\label{sec:hyper_param_details}
In~\name, the objective function (Eqn.~\ref{eqn:loss}) utilizes positive pairs of nodes co-occurring in fixed-length random walks. We follow the strategy of Deepwalk~\citep{deepwalk} to sample walks 10 walks of length 40 per node, each with a context window size of 10. 
We use 10 negative samples per positive pair, with context distribution ($P^t_n$) smoothing over node degrees with a smoothing parameter of 0.75, following~\citep{deepwalk,node2vec,graphsage}.
During training, we apply $L_2$ regularization with $\lambda  = 5 \times 10^{-4}$ and use dropout rates~\citep{dropout} of 0.1 and 0.5 in the self-attention layers of the structural and temporal blocks respectively. We use the validation set for tuning the learning rate in the range of $\{ 10^{-4}, 10^{-3}\}$ and negative sampling ratio $w_n$ in the range $\{ 0.01, 0.1, 1\}$.

We tune the hyper-parameters of all baselines following their recommended guidelines.
For node2vec, we use the default settings as in the paper, with 10 random walks of length 80 per node and context window of 10, trained for a single epoch. We tune the in-out and return hyper-parameters, $p,q$ using grid-search, in the range $\{0.25, 0.50, 1, 2, 4\}$ and report the best results. In case of GraphSAGE, we train a two layer model with respective neighborhood sample sizes 25 and 10, for 10 epochs, as described in the original paper. We evaluate the embeddings at each epoch on the validation set, and choose the best for final evaluation. Note that the results of GraphSAGE reported in Table~\ref{tab:results} represent that of best-performing aggregator in each dataset.

DynamicTriad~\citep{dynamictriad} was tuned using their two key hyper-parameters determining the effect of smoothness and triadic closure,  $\beta_0$ and $\beta_1$ in the range $\{0.01,  0.1, 1, 10\}$, as advised, while using recommended settings otherwise. We use the $L_1$ operator ($|\ve^t_u - \ve^t_u|$) instead of Hadamard, as recommended in the paper, which also gives better performance.
For DynGEM, we tune the different scaling and regularization hyper-parameters, $\alpha \in \{ 10^{-6}, 10^{-5}\}$, $\beta \in \{ 0.1, 1, 2, 5\}$, $\nu_1 \in \{10^{-6},  10^{-4}\}$ and $\nu_2 \in \{10^{-6},  10^{-4} \}$, while using other default configurations. 
DynAERNN was tuned using similar scaling and regularization hyper-parameters, $\beta \in \{ 0.1, 1, 2, 5\}$, $\nu_1 \in \{10^{-6},  10^{-4}\}$ and $\nu_2 \in \{10^{-6},  10^{-4} \}$, while using similar layer configurations as~\name~to ensure direct comparability.

\section{Additional Dataset Details}
\label{sec:dataset_details}
In this section, we provide some additional, relevant dataset details. Since dynamic graphs often contain continuous timestamps, we split the data into multiple snapshots using suitable time-windows such each snapshot has an equitable yet reasonable number of interactions (communication/ratings). In each snapshot, the weight of a link is determined by the number of interactions between the corresponding pair of users during that time-period. The pre-processed versions of all datasets will be made publicly available, along with the scripts used for processing the raw data.

\textbf{Communication Networks.}
The original un-processed Enron dataset is available at \url{https://www.cs.cmu.edu/~./enron/}. We use only the email communcations that are between Enron employees, \textit{i.e.}, sent by an Enron employee and have at least one recipient who is an Enron employee. A time-window of 2 months is used to construct 16 snapshots, where the first 5 are used as warm-up (due to sparsity) and the remaining 11 snapshots for evaluation. 

The UCI dataset was downloaded from \url{http://networkrepository.com/opsahl_ucsocial.php}. This dataset contains private messages sent between users over a span of six months, on an online social network platform at the University of California, Irvine. The snapshots are created using their communication history with a time-window of 10 days. We discard/merge the terminal snapshots if they do not contain sufficient communications. 

\textbf{Rating Networks.}
We use the Round 11 version of the Yelp Dataset Challenge~\url{https://www.yelp.com/dataset/challenge}. To extract a cohesive subset of user-business ratings, we first select all businesses in the state of Arizona (the state with the largest number of ratings) with a selected set of restaurant categories. Further, we filter the data to retain only users and business which have at-least 15 ratings. Finally, we use a time-window of 6 months to extract 12 snapshots in the period of 2009 to 2015.

The ML-10m dynamic user-tag interaction network was downloaded from~\url{http://networkrepository.com/ia-movielens-user2tags-10m.php}. This dataset depicts the tagging behavior of MovieLens users, with the tags applied by a user on her rated movies. We use a time-window of 3 months to extract 13 snaphots over the course of 3 years.

\end{document}